\documentclass[preprint, 12pt]{elsarticle}
\usepackage{lipsum}
\makeatletter
\def\ps@pprintTitle{%
 \let\@oddhead\@empty
 \let\@evenhead\@empty
 \def\@oddfoot{}%
 \let\@evenfoot\@oddfoot}
\makeatother

\usepackage{graphicx}
\usepackage{pgfplots}

\usepackage{mathrsfs}
\usepackage{amsmath}

\usepackage{amssymb}

\usepackage{lineno}

\usepackage{multirow}

\usepackage{array}
\usepackage{subcaption}
\usepackage{float}

\newcolumntype{L}[1]{>{\raggedright\let\newline\\\arraybackslash\hspace{0pt}}m{#1}}
\newcolumntype{C}[1]{>{\centering\let\newline\\\arraybackslash\hspace{0pt}}m{#1}}
\newcolumntype{R}[1]{>{\raggedleft\let\newline\\\arraybackslash\hspace{0pt}}m{#1}}

\newcommand*{\Prob}{\mathsf{P}}
\newcommand{\softmax}{\mathop{\rm softmax}}
\newcommand{\TT}{\mathop{\rm TT}}

\journal{}

\begin{document}

\begin{frontmatter}

\title{Compression of Recurrent Neural Networks for Efficient Language Modeling}

\author[inst1,inst2]{Artem~M.~Grachev}
\ead{grachev.art@gmail.com}
\author[inst2]{Dmitry~I. Ignatov}
\author[inst3]{ Andrey~V. Savchenko}

\address[inst1]{Samsung R\&D Institute, Moscow, Russia}
\address[inst2]{National Research University Higher School of Economics, Moscow, Russia}
\address[inst3]{National Research University Higher School of Economics, Laboratory of Algorithms and Technologies for Network Analysis, Nizhny Novgorod, Russia}

\begin{abstract}

Recurrent neural networks have proved to be an effective method for statistical language modeling. However, in practice their memory and run-time complexity are usually too large to be implemented in real-time offline mobile applications.
In this paper we consider several compression techniques for recurrent neural networks including Long-Short Term Memory models. We make particular attention to the high-dimensional output problem caused by the very large vocabulary size. 
We focus on effective compression methods in the context of their exploitation on devices: pruning, quantization, and matrix decomposition approaches (low-rank factorization and tensor train decomposition, in particular). For each model we investigate the trade-off between its size, suitability for fast inference and perplexity. We propose a general pipeline for applying the most suitable methods to compress recurrent neural networks for language modeling.  It has been shown in  the experimental study with the Penn Treebank (PTB) dataset that the most efficient results in terms of speed and compression-perplexity balance are obtained by matrix decomposition techniques.

\end{abstract}


\end{frontmatter}

%\linenumbers

\section{Introduction}

With the increasing popularity of neural networks, the question of their implementation on mobile devices is actually emerging.
Consider the language modeling problem~\cite{croft2003language, deng2018feature}, in which it is required to develop a probabilistic mechanism for generating text (statistical language model). This problem appears in many practical applications, i.e., text sequence generation, machine translation, speech recognition and text analysis. 
The first methods to create a language model were based on the storing of all possible continuations for a given beginning sequence of words. Those methods are the so-called $n$-gram models~\cite{JelMer80, Kneser:ibfmlm:1995,Jelinek:SMfSR:1997}. %вставить ссылки
Unfortunately, it is known that these models have common issues. 
For example, all the chains of the length $n=4$ for a fixed vocabulary $\mathbb{V}$  of size $|\mathbb{V}| = 10,000$ naturally occupy several gigabytes and the taken memory grows exponentially with an increase of $n$. 
Thus, the chains of length $n=20$ cannot be physically stored in the memory of modern computers.

Recurrent neural networks (RNN) can solve this problem in certain aspects. It was shown 
that contemporary Long-Short Term Memory (LSTM) or Gated Recurrent Units (GRU) models can take into account the long-term dependencies between words~\cite{Schmid:LSTM:1997,GRU:Cho:2014}. 
Nowadays RNN-based models are implemented in various practical tasks of natural language processing due to their ability to provide high accuracy and can be robustly trained by using the well-established hyperparameters. Though conventional convolutional neural networks \cite{DBLP:conf/emnlp/PhamKB16, DBLP:conf/aaai/KimJSR16} can be also used in these tasks, they are limited by a fixed-length context and cannot directly learn longer-term dependencies. The modern transformer models \cite{DBLP:journals/corr/abs-1810-04805, DBLP:conf/nips/VaswaniSPUJGKP17} based on attention mechanism are not still well-studied due to their difficult training process. Hence, in this paper we decided to focus on modern RNN-based language models.

Unfortunately, such language models still have large memory complexity. Hence, they are inappropriate for use in embedded systems, e.g., mobile phones, which usually have less computational resources for modern applications (for example see in \cite{cheok2008efficient}) when compared to modern graphical processor units (GPU). This problem is especially challenging for the language modeling task, in which the RNNs are characterized by very large dimensionality of the last fully connected layer~\cite{Bengio+Senecal-2003} because  this layer produces $|\mathbb{V}| \gg 1$ posterior probabilities of all words from the vocabulary.

The main contribution of this paper is as follows. Firstly, we proposed a pipeline to compress the recurrent neural networks for language modeling. Such networks can be roughly structured as architectures containing the input embedding level for  continuous representation of words in the vector space, recurrent cells and the output layer for prediction of the next word in a sequence. In the paper we compress the output layer and LSTM-layers separately. We mainly focus on the matrix factorization compression techniques resulting in low-rank matrices as well as more complex approaches like Tensor-Train decomposition. 
Secondly, we made particular attention to the high-dimensional output problem caused by the very large vocabulary size. This problem is especially challenging in the language modeling tasks with word prediction. We presented the solution of this problem using low-rank and Tensor Train decomposition of the weight matrices in the last fully-connected layer of the neural network.
Thirdly, we implemented our models for GPUs of a mobile phone in order to test the studied algorithms on real devices. It was experimentally shown that compression techniques based on matrix factorization sufficiently speed up the inference in the language model. 

This article is an extended version of our conference paper~\cite{Grachev1}. In comparison with our previous paper we: 1) formulated the methodology for compression of language models; 2) presented an approach to solving the high-dimensional output problem; 3) significantly extended the survey of related works and references section; 4) provided plenty of new experiments for compression of conventional baselines (i.e., those that were described by Zaremba et al~\cite{Zaremba:RNNReg:2014}) including their several extensions; 5) measured inference time for GPU computations on a real mobile device.

The paper is organized as follows. Section~\ref{sec:relwork} overviews related works. In Section~\ref{sec:languagem}, we give an overview of language modeling task and then focus on respective RNN-based approaches to this task. Next, in Section~\ref{compression_methods}, we describe different types of compression. In Subsection~\ref{sec:prunquant}, we consider the simplest known methods for neural networks compression like pruning and quantization. In Subsection~\ref{sec:lr}, we consider compression of neural networks based on different matrix factorization methods. Subsection~\ref{sec:tt} deals with Tensor Train decomposition (TT-decomposition).
In Subsection~\ref{sec:general_pipeline} we present the general pipeline of the methodology to compress RNN-based language models and make them suitable for offline usage in mobile devices. 
Section~\ref{results} describes our experimental results and important implementation details.
Finally, in Section \ref{sec:conclusion}, we summarize the results of our work and discuss future research.

\section{Related works}\label{sec:relwork}

Generally, there are several approaches to the neural network compression.
They can be roughly divided into two kinds, namely, the methods based on sparse computations and 
the methods based on using different properties of weights matrices, e.g., matrix factorization. 

The first kind of techniques include pruning and quantization and was originally applied in computer vision. 
In one of the first works on these methods~\cite{pruning1} it was shown that pruning makes it possible to remove a lot of weights before doing quantization without loss of accuracy. It was verified for such neural networks as LeNet, AlexNet, VGGNet that pruning can remove $67\%$ for convolutional layers and up to $90\%$ for fully connected layers. Moreover, an even higher compression ratio can be achieved by combining pruning with mixed precision~\cite{Han:DeepCompressing:2016}. 

The pruning techniques have been justified in terms of variational dropout. The variational dropout was introduced in~\cite{vardropout} as a method for automatic tuning of a proper dropout rate, i.e. the probability that a given neuron will become active. 
In \cite{bayescomperssion1,bayescomperssion2}, the authors adapt this dropout techniques for the neural network compression. 
In their studies, the dropout rate is allowed to be equal to one, which is equivalent to a complete elimination of this neuron from a network. 
As a matter of fact, pruning and quantization are able to provide a rather large reduction of the size of a trained network for the models stored on a hard disk. 
However, there are several issues when we try to use such models in the inference phase. 
They are caused by high computation time of sparse computing with the prunned matrices. The example of one possible solution is the so-called structured pruning as for example 
in~\cite{bayescomperssion2, bayescomperssion3}, when a set of rows or columns is dropped in a certain layer matrix. 

Another branch of compression methods include different matrix decomposition approaches by using either matrices of lower sizes or exploiting the properties of special matrices involved in a compression method. For example, the paper~\cite{Bengio:URNN:2016} proposes a new type of RNNs based on potentially more compact unitary matrices, called Unitary Evolution Recurrent Neural Networks (better than RNN in copying memory and adding problems).
In \cite{Yang:2015}, the authors applied the so-called FastFood transform for fully-connected and convolutional layers (up to 90\% of compression in terms of the number of parameters stored).
Different matrix decomposition techniques can be also related to this class of compression methods. 
These methods can be as simple as low-rank decomposition or more complex like Tensor Train (TT)
decomposition~\cite{novikov15tensornet,garipov16ttconv,TTRNN,yu2017long}. However, the TT-based approach have not been studied in language modeling task, where there are such issues as high-dimensional input and output data, and, as a consequence, more options to configure TT decomposition.
Thus, the second kind of methods allows compressing neural networks, get reduction in the model size and still have suitable non-sparse matrices for multiplication. 

Most of described methods can be used for compression of RNNs. Methods based on matrix decomposition of RNNs were mainly applied in automatic speech recognition \cite{Lu:LR:2016, DBLP:conf/icassp/PrabhavalkarABM16, TTRNN}. For example, the usage of the Toeplitz-like structured matrices in~\cite{Lu:LR:2016} gave up to 40\% compression of RNN for voice search task. Compression of embeddings layers is considered in the paper~\cite{DBLP:journals/corr/abs-1811-00641}. 

One of the most challenging problems, which influenced the large size of a language model, is the very-high dimensionality at the output layer caused by the huge size of vocabulary. This problem is widely discussed in the literature. 
For example, peculiarities of unknown and rare words presence are considered in~\cite{GulcehreANZB16}.
Huge computation complexity and big size of the softmax layer is discussed in~\cite{Bengio+Senecal-2003,Morin05hierarchicalprobabilistic}. Morin et al~\cite{Morin05hierarchicalprobabilistic} develop the idea of hierarchical softmax computation. They show that it is possible to obtain $O(\sqrt{|\mathbb{V}|})$ parameters for this way of softmax computation. Bengio et al.\cite{Bengio+Senecal-2003} propose the method for speeding up softmax computation by sampling of subset of words from the available vocabulary on each iteration during the training phase.  

The pruning with the variational dropout technique was applied to compression of RNNs in natural language processing tasks \cite{bayescomperssion3, DBLP:conf/emnlp/ChirkovaLV18}. However, the results in language modeling~\cite{bayescomperssion3} are significantly worth in terms of achieved perplexity when compared even with classical results of Zaremba et al~\cite{Zaremba:RNNReg:2014}. Moreover, the acute problem with high-dimensional output is completely ignored.

Hence, it seems that there is still no study which provides a methodology to compress RNNs in language modeling, which overcomes the difficulties peculiar to high-dimensional output layers and makes it possible to achieve both low memory and run-time complexity in order to be implemented in real-time offline mobile applications.

\section{RNNs in language modeling problem}\label{sec:languagem}

Consider the language modeling problem, in which it is required to estimate the probability of a sentence or
sequence of words $(w_1, \ldots, w_T)$
in a language $L$.
\begin{multline}
  \Prob{\left(w_1, \ldots, w_T\right)} = 
  \Prob{\left(w_1, \ldots, w_{T-1}\right)}
  \Prob{\left(w_T | w_1, \ldots, w_{T-1}\right)} = \\
  = \prod_{t = 1}^{T}\Prob{\left(w_t | w_1, \ldots, w_{t-1}\right)}
\end{multline}

The use of such a model directly requires estimation of posterior probability
$\Prob{\left(w_t | w_1, \ldots, w_{t-1}\right)}$. In general, this estimation has too much run-time complexity.
Hence, a typical approach approximates it with the probability $\Prob{\left(w_t | w_{t-n}, \ldots, w_{t-1}\right)}$ of the next word given a fixed number $n$ of previous words.
This naturally leads us to $n$-gram models~\cite{Kneser:ibfmlm:1995, Jelinek:SMfSR:1997}, in which a discrete probability distribution $\Prob{\left(w_t | w_{t-n}, \ldots, w_{t-1}\right)}$ is given by a table with $(n+1)$ columns, which contains the count of phrases with $(n+1)$ sequential words in a large text corpora. It was a common approach for language modeling until the middle of the 2000s. Unfortunately, such an approach requires a very large memory to store the long term dependencies. Moreover, the probabilities of rare phrases are usually underestimated.

Thus, a new milestone in the domain had become the use of RNNs, which were successfully implemented for language modeling in the papers~\cite{DBLP:conf/interspeech/MikolovKBCK10, Bengio03aneural, Mikolov:2007}. Consider an RNN, where $L$ is the number of recurrent layers, $x_{\ell}^{t}$ is the input of the layer $\ell$ at the moment $t$. 
Here $t \in \{1,\ldots,T\}$, $\ell \in \{1, \ldots, L\}$, and $x_{0}^{t}$ is the embedding vector. We can describe each layer as follows: 

\begin{align}
  \label{RNN_one_layer} z_{\ell}^t = & W_{\ell}x_{\ell-1}^{t} + U_{\ell}x_{\ell}^{t-1} + b_l   \\
    x_{\ell}^t = & \sigma(z_{\ell}^t),
\end{align}
where $W_{\ell}$ and $V_{\ell}$ are matrices of weights and $\sigma$ is an activation function. %Usually ReLu. 
The output of the network uses the softmax activation:
\begin{equation}
    \label{output_layer}
    y^t = \softmax\left[W_{L+1}x_{L}^t + b_{L+1}\right].
\end{equation}
Then, we estimate the posterior probability in Eq. 1) as an output of such RNN:
\begin{gather}
\Prob{\left(w_t | w_1, \ldots, w_{t-1}\right)} = y^t.
\end{gather}

While the $n$-gram models 
even with not very large $n$ require a lot of memory space due to the combinatorial explosion, RNNs can learn representations of words and their sequences without memorizing directly all word contexts.

Nowadays the mainly used variations of RNN are designed to solve the problem of vanishing gradients, which usually appears in the training with long sequences
~\cite{deng2018feature, Gradient_flow:Hochreiter:2001}. The most popular implementations of the RNNs which do not suffer from this problem are LSTM~\cite{Schmid:LSTM:1997} and GRU~\cite{GRU:Cho:2014} networks. 

RNN-based approaches to the language modeling problem are efficient and widely adopted, but still require a lot of space.
For example, each LSTM layer with the input dimensionality $k$ and output size $k$ involves eight matrices of size  $k\times k$. Moreover, usually in language modeling applications, one wants the model to use words (rather than characters) as the fundamental units as the input and the output. These peculiarities naturally lead us to large sizes for both the input and output layers. 
The input layer is an embedding layer that maps every word from vocabulary $\mathbb{V}$ to a vector. The output layer is an affine transformation from a hidden representation to the output space, for which then we apply the softmax function. The size of vocabulary, $|\mathbb{V}|$, is of the order of thousands or even tens of thousand. 
Hence, the matrices in the input and output layers contain $|\mathbb{V}|\times k$ parameters. Thus, the number of parameters in the whole network with $L$ LSTM layers and the dimensionality of the input embeddings identical to the size of the hidden state $k$ is given by
\begin{equation}
\label{nparam}
n_{total} = 8Lk^2 + 2|\mathbb{V}|k .
\end{equation}

Let us analyze the contribution of each term in (\ref{nparam}) in a realistic scenario. 
For the PTB dataset we have the vocabulary size $|\mathbb{V}|  = 10,000$. 
 Consider an LSTM network with two hidden layers of $k=650$ units in each layer. Each LSTM layer includes eight matrices of size
 $650\times 650$, i.e.
 $650 \times 650 \times8 \times 2 = 6.76 \text{M parameters}.$
 The output layer in this network have 
 $650 \times 10000 = 6.5 \text{M parameters}$. 
 Similar calculations for the LSTM with the size of hidden layers of 1500 give us 
 36M parameters
 and 
 15M parameters, respectively.
Thus, we can see that the output (softmax) layer can occupy up to one third of the whole network.
Note that the embedding of a network can occupy the same memory size if we do not use techniques like ``tied softmax''~\cite{DBLP:journals/corr/InanKS16, E17-2025}. Hence, in this paper we decided to address this problem by performing several experiments with the softmax layer using low-rank decomposition and TT decomposition to reduce its size.

\section{Compression methods}
\label{compression_methods}

\subsection{Pruning and quantization}\label{sec:prunquant}

In this subsection, we consider very simple though not the most effective techniques to compress neural networks. 
Some of them were successfully applied to audio processing~
\cite{Han:DeepCompressing:2016} and image processing~\cite{Image:Svachenko:2017}. However, they are not yet well-studied in the language modeling task~\cite{Grachev1}.

\begin{figure}[h]
\begin{tikzpicture}
\pgfplotstableset{col sep = comma}
    \begin{axis}[width=0.95\textwidth, height=6.5cm, xmin=-3.5, xmax=3.5, ymin=0, ymax=180000, xlabel={\footnotesize Value}, ylabel={\footnotesize Frequency},
                x tick label style={font=\footnotesize}, y tick label style={font=\footnotesize}, area style
          ]
         \addplot[ybar interval,mark=no] table {pics/left.csv};\label{gr:before_pr}

    \end{axis}
\end{tikzpicture}

\begin{tikzpicture}
\pgfplotstableset{col sep = comma}
    \begin{axis}[width=0.95\textwidth, height=6.5cm, xmin=-3.5, xmax=3.5, xlabel={\footnotesize Value}, ylabel={\footnotesize Frequency},
                x tick label style={font=\footnotesize}, y tick label style={font=\footnotesize}, area style
          ]
         \addplot[ybar interval,mark=no] table {pics/right.csv};\label{gr:after_pr}

    \end{axis}
\end{tikzpicture}

\caption{An example of weights distribution before (top) and after pruning (bottom)
 }\label{fig:prun}
\end{figure}
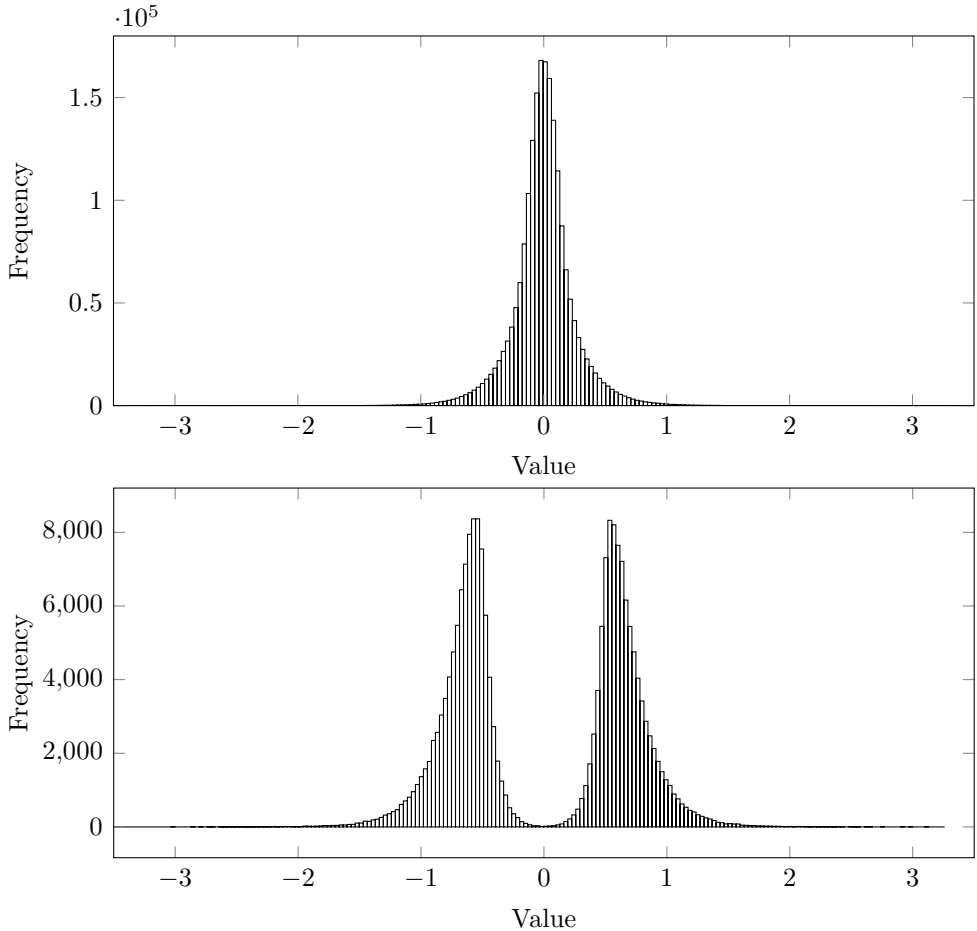

Pruning is a method for reducing the number of parameters of a neural network by removing the weights, which are approximately equal to zero. In Fig.~\ref{fig:prun}~(top), one can notice that  
usually the majority of weight values is concentrated near zero. It means that such weights do not provide a valuable contribution to the final output. Hence, we can remove from the network all the connections with the weights, which do not exceed a certain threshold (Fig.~\ref{fig:prun}~(bottom)). After that, the network is fine-tuned to learn the final weights for the remaining sparse connections.

Quantization is a method for reducing the size of a compressed neural network in memory. In this technique, each floating-point value of weight is packed into, e.g., 8-bit integer representing the closest real number in one of 256 equally-sized intervals within the whole range of the weight.

Pruning and quantization have common disadvantages since they do not support training from scratch. Moreover, their practical usage is
quite laborious.  
The reason for such behaviour of pruning mostly lies in the inefficiency of sparse computing.
In the case of quantization, the model is stored in an 8-bit representation, 
but 32-bits computations are still required. It means that we do not obtain advantages using in-memory techniques at least until  the tensor processing unit (TPU) is not used,
which is adapted for effective 8- and 16-bits computations.

\subsection{Low-rank factorization}\label{sec:lr}

Low-rank (LR) factorization~\cite{Lu:LR:2016} represents more powerful techniques to compress the matrices of weights. A simple LR decomposition for RNN can be implemented as follows:

\begin{equation}
    x_l^t = \sigma\left[W_\ell^aW_\ell^bx_{\ell-1}^t + 
    U_l^aU_l^bx_{\ell}^{t-1} + b_l\right]
\end{equation}
%\vspace
The paper~\cite{Lu:LR:2016} requires the following constraint: $W_l^b = U_{\ell-1}^b$. Thus, the RNN equation can be  rewritten as follows: 
\begin{align}
    x_l^t & =   \sigma\left[W_l^a m_{l-1}^t + U_{l}^a m_{l}^{t-1} + b_l \right] \\
    m_l^t &  =   U_l^b x_l^t
    \\
    y_t & =  \softmax\left[W_{L+1}m_L^t + b_{L+1}\right]
\end{align}

Compression of LSTM and GRU layers is implemented in a similar way but with slightly more complex equations. First let us describe one layer of LSTM: 

\begin{align}
\label{eq:input_gate}
    i_{\ell}^t = & \: \sigma\left[W_l^i x_{l-1}^{t} + U_l^i x_{l}^{t-1} + b_{l}^{i}\right] 
    & \text{input gate}  \\ 
\label{eq:forget_gate}
    f_{\ell}^t = & \: \sigma\left[W_l^f x_{l-1}^{t} + U_l^f x_{l}^{t-1} + b_{l}^{f}\right] 
     & \text{forget gate}  \\
\label{eq:cell_eq}
    c_{\ell}^t = & \: f_l^t \odot c_l^{t-1} + i_l^t \tanh\left[W_l^c x_{l-1}^t + U_l^c x_l^{t-1} + b_l^c\right] 
     & \text{cell state} \\
\label{eq:output_gate}
    o_{\ell}^t = & \: \sigma\left[W_l^o x_{\ell-1}^{t} + U_l^o x_{l}^{t-1} + b_{l}^{o}\right] 
     & \text{output gate}  \\ 
\label{eq:lstm_eq3}
    x_{\ell}^t = & \: o_{\ell}^t \cdot \tanh[c_l^t],
\end{align}

And here are equations for the GRU layer:

\begin{align}
\label{gru:zt}
z_l^t = & \: \sigma\left( W_l^z x_{l-1}^t + U_l^z x_l^{t-1}\right) & \text{update gate}
\\\label{gru:rt} 
r_l^t = & \: \sigma\left(W_l^r x_{l-1}^t  + U_l^r x_{t-1}\right) & \text{reset gate} \\
\label{gru:proposal}
\tilde{x}_l^t = & \: \tanh \left(W_l^h x_{l-1}^t  + U_l^h \left(r_l^t \cdot x_l^{t-1}\right) \right) & \text{proposal output} \\
\label{gru:final}
x_t^l = & (1 - z_l^t)\odot x_{l-1}^t  + z_l^t \odot \tilde{x}_l^t & \text {final output}
\end{align}
where $c_{\ell}^t$ is the memory vector at the layer $\ell$ and time step $t$. The output of the network is given by the same Eq.~\ref{output_layer} as above.

The compressed LSTM is described by the same equations~\ref{eq:input_gate}-\ref{eq:lstm_eq3}, but the sizes of matrices are changed. Here, similarly to the RNN case, 
we require existence of a special matrix $W_l^p$ such that $W_l^{ib} = W_l^{fb} = W_l^{cb}=W_l^{ob} 
= U_{l-1}^{ib} = U_{l-1}^{fb} = U_{l-1}^{cb}=U_{l-1}^{ob} = W_l^p $
and the output of the network is computed as follows: \begin{equation}
\hat{x}_l^l = W_l^p x_l^t.
\end{equation}

The situation is more peculiar with the GRU cells. If we just reduce the sizes of matrices $W_l^z, U_l^z, W_l^f, U_l^f$ to $k \times r$, we will end up with wrong dimensions in Eq.~
(\ref{gru:proposal}-\ref{gru:final}). That is why we 
reduce those matrices down to $r \times r$, reduce $W_l^h, U_l^h$ down to $r \times k$ and introduce the projection matrix $W_l^p \in \mathbb{R}^{k \times r}$ after Eq.~\ref{gru:proposal} so that Eq.~\ref{gru:final} is applied in the same way by replacing $\tilde{x}_t^l$ to:

\begin{equation}
x_t^{lp} = W_l^p \tilde{x}_t^l.
\end{equation}

The main advantage of the LR technique lies in potentially small sizes $r \times k$ and $k \times r$ of matrices $W_l^a$/$U_l^b$ and $U_l^a$, respectively (in case of RNN). Those sizes are much less than the size $k \times k$ of the original weight matrices, $W_l$ and $V_l$, if $r \ll k$. With a reasonably small $r$ we obtain the advantage both in size and multiplication speed. The same considerations 
are valid for LSTM and GRU cells with 8 and 6 matrices, respectively.

\subsection{Tensor Train decomposition}\label{sec:tt}

Taking into account the recent advances of TT decomposition in deep learning~\cite{novikov15tensornet,garipov16ttconv}, we have also decided to apply this technique to recurrent neural network compression for language modeling.

The TT decomposition was originally proposed as an alternative and more efficient form of tensor representation~\cite{TT:Oseledets:11}. Let us describe how this decomposition could be applied to neural networks. Consider, for example, the weights matrix $W \in \mathbb{R}^{k \times k}$ of the RNN layer (\ref{RNN_one_layer}). One can arbitrarily choose such numbers $k_1, \ldots, k_d$ so that $k_1 \times \ldots \times k_d = k \times k$, and reshape the weights matrix to a tensor $\vec{W} \in \mathbb{R}^{k_1 \times \ldots \times k_d}$. Here $d$ is an order (degree) of a tensor, $k_1, \ldots, k_d$ are the sizes of each dimension. Thus we can perform the TT-decomposition of the tensor $\vec{W}$ and obtain a set of matrices $G_m[i_m] \in  \mathbb{R}^{r_{m-1} \times r_m}, 
i_m = 1, \ldots, k_m$, $m = 1,\ldots, d$ and $r_0 = r_d = 1$ such that each of the tensor element can be represented as $\vec{W}(i_1, i_2, \ldots, i_d) = G_1[i_1]G_2[i_2]\ldots G_d[i_d]$. Here $r_{0},\ldots r_{m}$ are the ranks of the decomposition.
Such TT decomposition can be efficiently implemented with the TT-SVD algorithm described in~\cite{TT:Oseledets:11}. 
In fact, each $G_m  \in  \mathbb{R}^{r_{m-1} \times k_m \times r_m} $ is a three-dimensional tensor with the second dimension $k_m$ corresponding to the  dimension of the original tensor and two ranks $r_{m-1}, r_m$, that in certain sense is a size of an internal representation for this dimension. It is necessary to emphasize that even with the fixed number for dimensions of reshaped tensors and their sizes we still have plenty of variants to choose the ranks in the TT-decomposition.

Let us denote these two operations of converting matrix $W$ to $\vec{W}$ and decomposing it on TT format as one operation $\TT(W)$. Applying it to both the matrices $W$ and $V$ from Eq.~\ref{RNN_one_layer} we obtain TT-RNN layer in next form:  

\begin{equation}
   z_{\ell}^t = \sigma( \TT(W_l)x_{\ell-1}^{t} + \TT(U_l)x_{\ell}^{t-1} + b_{\ell}).
\end{equation}

Similarly we can apply TT-decomposition to each matrix of LSTM layer (\ref{eq:input_gate})-(\ref{eq:output_gate}) or the matrix of the output layer (\ref{output_layer}). Moreover, according to~\cite{TT:Oseledets:11}, the matrix-by-vector product and matrix sum can be efficiently implemented directly in the TT format without the need to convert these matrices to the TT. 

The TT compression can be achieved by choosing the internal ranks $r_1, \ldots, r_{d-1}$.  Let $R = \max\limits_{m = 0,\ldots, d} r_m$, $K = \max\limits_{m = 0,\ldots, d} k_m$. Hence, the number of parameters for the TT-decomposition is equal to $N_{\TT} = \sum r_{m-1}k_m r_{m} \leq dR^2K$. In fact, each factor in this product can be smaller an order of magnitude than the original $k$. 

This approach was successfully applied to compress fully connected neural networks~\cite{novikov15tensornet}, to develop convolutional TT layer~\cite{garipov16ttconv} and to compress and improve RNNs~\cite{TTRNN,yu2017long}. However, there are still no studies of the TT decomposition for language modeling and similar tasks with high-dimensional outputs at the softmax layer.

\subsection{Proposed pipeline for compressing RNN model} 
\label{sec:general_pipeline}

\begin{figure}[h]
\centering
   \includegraphics[width=0.95\textwidth]{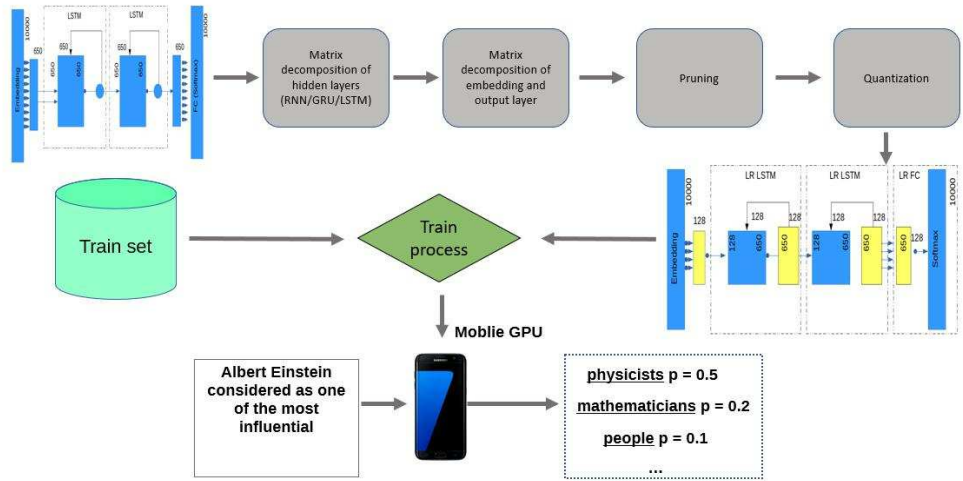}
   \caption{Proposed pipeline of the RNN compression for efficient language modeling}
   \label{fig:pipeline} 
\end{figure}

To sum it all up, we propose a general pipeline (Fig.~\ref{fig:pipeline}) for the compression of RNNs in the language modeling problem.  
Here, firstly, the internal recurrent layers of RNNs are compressed. Then we continue by compression of external embedding layers and the output one by using either conventional matrix factorization or TT decomposition. In addition, pruning and quantization can be applied for the preliminary decomposed neural nets. The resulted compressed language model can be used directly on mobile devices. We developed an optimized version of inference in the LR-factorized RNNs using the GPU of a modern mobile phone.

 %Сжатые архитектуры мы протестировали на мобильном устройстве. Была реализована оптимизированная версия сжатых нейронных сетей для мобильного телефона. 

\section{Experimental results}\label{results}

\subsection{Experimental setup}

We choose the models from~\cite{Zaremba:RNNReg:2014} as a reliable baseline. By default, there are Small, Medium, and Large LSTM models with the sizes of their hidden layers 200, 650, and 1500, respectively, but we provide additional experiments with different hidden sizes and with different types of cells like RNN and GRU. All experiments are produced with the PTB (Penn TreeBank) dataset~\cite{DBLP:conf/interspeech/MikolovKBCK10}. 

We compare all the models in terms of two quality metrics: \textbf{perplexity} and \textbf{the number of parameters}. 
The perplexity of language models is a conventional quality metric for language modeling. The value $p$ of perplexity shows that the model is as confused on test data as if it had to choose uniformly and independently among $p$ options for each word. In addition, we characterize the quality of the model by the \textbf{average word prediction accuracy}, i.e., the probability to correctly predict the next word in a sequence, which can be estimated as one divide by perplexity. 

Finally, we measure the average \textbf{inference time} using our own implementation of inference in RNNs for mobile GPUs. We have performed testing on a real mobile device to compare the performance of compressed models. A mobile phone Samsung S7 Edge with GPU processor Mali-T880 was used in our experiments.
The calculations are carried out after the ``warming phase'' (100 preliminary calculation loops) to achieve a maximum computational performance of the mobile GPU. The inference time results are averaged over 1000 runs for each model.

\begin{figure}[h]
\centering
\begin{subfigure}[h]{0.55\textwidth}
   \includegraphics[width=95mm]{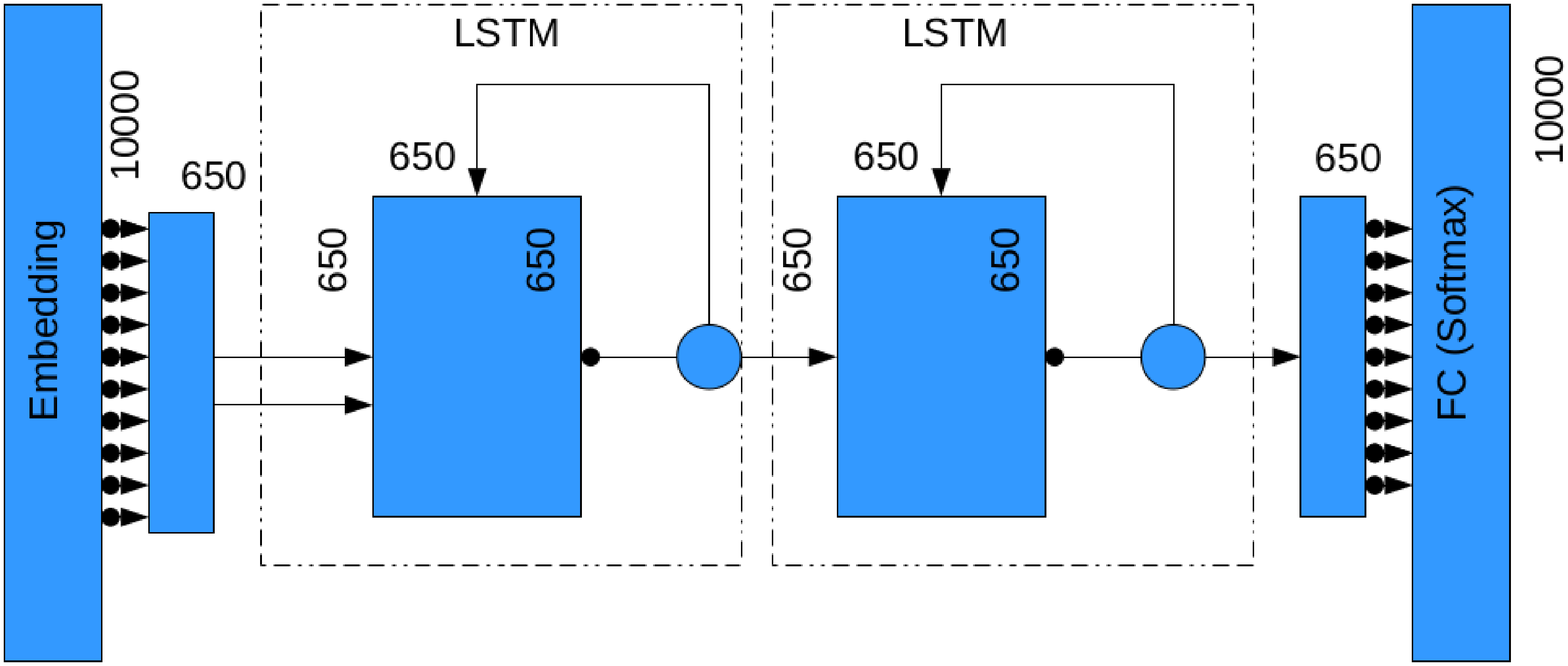}
   \caption{}
   \label{fig:Ng1} 
\end{subfigure}

\begin{subfigure}[h]{0.55\textwidth}
   \includegraphics[width=95mm]{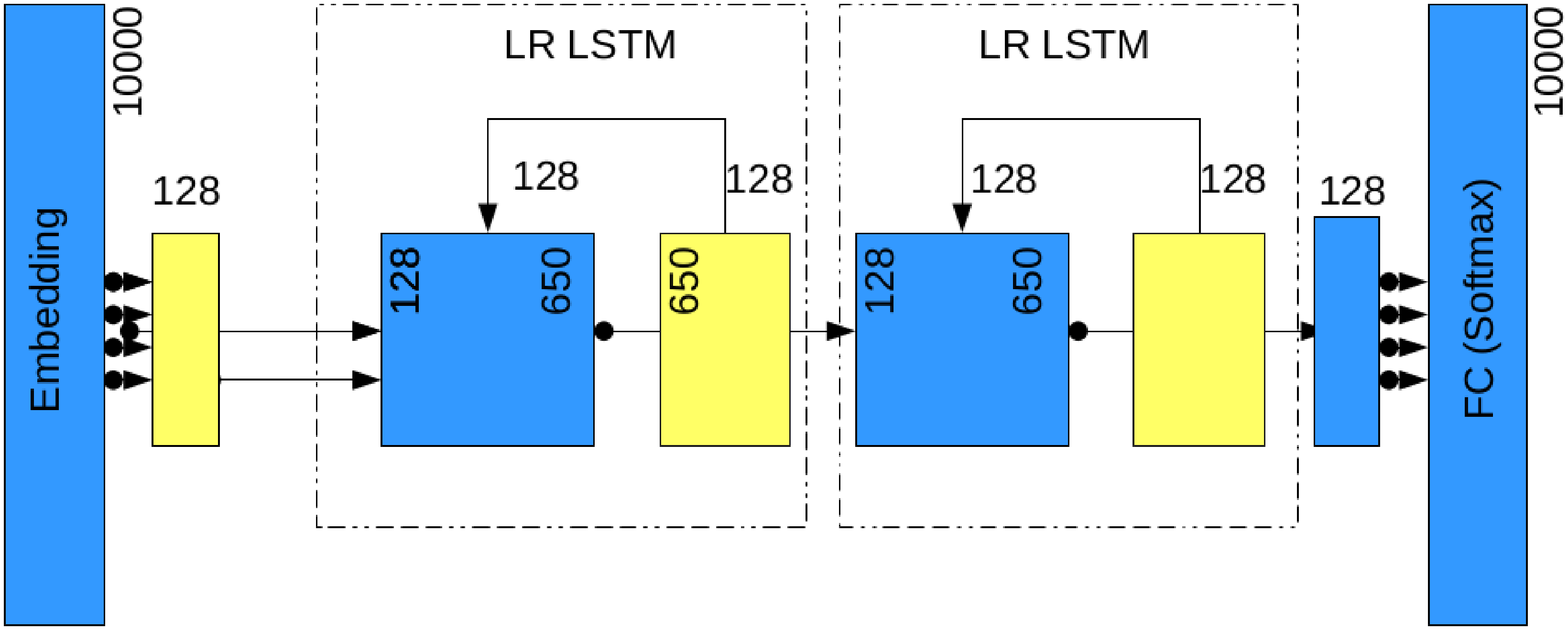}
   \caption{}
   \label{fig:Ng2}
\end{subfigure}
\caption{Neural network architectures: (a) original \textbf{LSTM~650-650}, (b) LR-compressed model}\label{fig:scheme}
\end{figure}

We implemented all the above-mentioned compression techniques. The pruning and quantization were tested for small LSTM model from \cite{Zaremba:RNNReg:2014}. In addition, we thoroughly studied how matrix factorization techniques perform for LSTM layers of different sizes as well as for nets based on such units as GRU and RNN. We have tried to hold compression ratio in a range x3-x5.  
%The matrix decomposition techniques were thoroughly studied in the experiments with Medium \textbf{(LSTM 650-650)} and Large \textbf{(LSTM 1500-1500)} LSTM models. 
For example, let us describe the sizes for one of the obtained decompositions with the \textbf{LSTM~650-650} model. 
We start with the initial sizes for $W \in \mathbb{R}^{650 \times 650}$ $U \in \mathbb{R}^{650 \times 650}$, and $|\mathbb{V}| = 10,000$. The corresponding matrix for the embedding is $W_{emb} \in \mathbb{R}^{10,000 \times 650} $ and the matrix for the output is $W_{out} \in \mathbb{R}^{10,000 \times 650}$.
The size of each weight matrix, $W$ and $U$, is reduced down to $650 \times 128$ and the sizes of the embedding and output matrices are down to $10,000 \times 128$ and $128 \times 10,000$, respectively. The value 128 is chosen as the most suitable degree of 2 for efficient device implementation. We have performed several experiments with other size options, but the above-mentioned configuration is the best in terms of compression-perplexity balance. The diagrams of the original and the LR-compressed models are shown in Fig.~\ref{fig:scheme}.

In order to appropriately choose the hyperparameters of our models and the training procedure (learning rate, schedule, dropout rate, sizes of decomposition, number of hidden neurons, etc.) and avoid expensive grid search in the space of the parameters, we have followed the random search procedure in the respective hyperparameter space. 

In the case of TT-decomposition under a fixed decomposition scheme, we examine different values of internal rank and choose the value, which provides the lowest perplexity. We set the basic configuration of an LSTM-network with two 600-600 layers and four tensors for each matrix in a layer. The size of layers is chosen as 600 by 600 instead of 650 by 650 due to better factorization for TT-decomposition with more divisors: $600 = 2\cdot 3\cdot 5 \cdot 2 \cdot 5 \cdot 2$ versus $650 = 2 \cdot 5 \cdot 5 \cdot 13$. Then we perform a grid search through a different number of dimensions and various tensor rank values. 

\begin{figure}[t!]

\begin{tikzpicture}
    \begin{axis}[width=\linewidth, height=8cm, xmin=0, xmax=95, xlabel={\footnotesize Epoch}, ylabel={\footnotesize Perplexity},
          xticklabels=  {0,5,10,15,20,25,30,35,40,45,50,55,60,65,70,75,80,85,90,95},
          xtick=        {0,5,10,15,20,25,30,35,40,45,50,55,60,65,70,75,80,85,90,95}, ymin=0, ymax=600,
          x tick label style={font=\footnotesize}, y tick label style={font=\footnotesize}
          ]
      \addplot[line width=.6pt,black,solid,mark=asterisk,mark repeat=1] table {pics/Train_Perplexity.txt};
      \addplot[line width=.6pt,black,solid,mark=o,mark repeat=1] table {pics/Valid_Perplexity.txt};
      \addlegendentry{Training}
      \addlegendentry{Validation}
    \end{axis}
\end{tikzpicture}
\vspace{-.5cm}
\caption{Learning curves for training \textbf{LR~LSTM~500-500} model}\label{fig:learn_curves}
\end{figure}
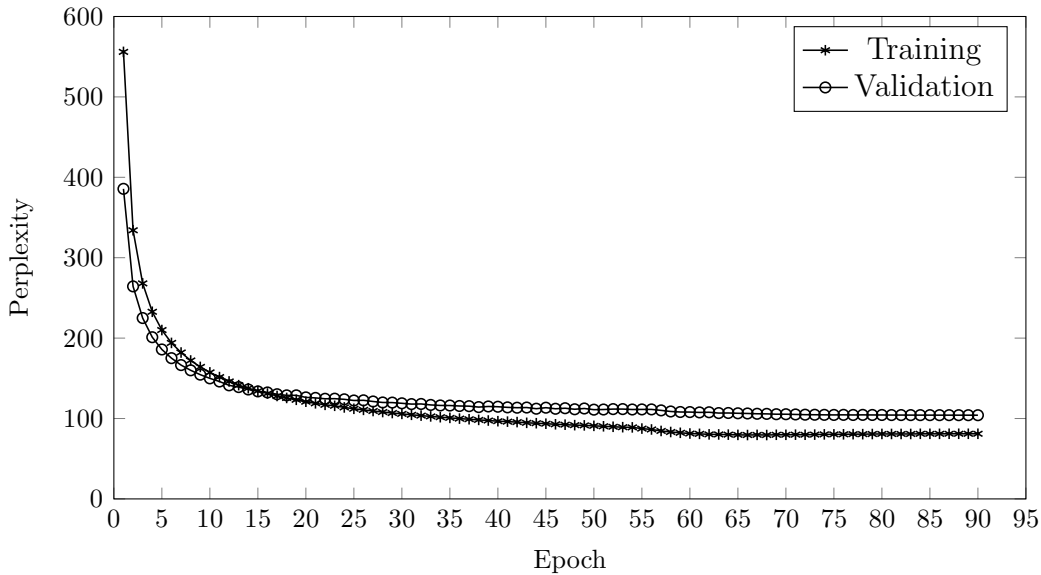

To obtain the best perplexity, we perform two stages of training for the LR-compressed models. At first, Adam optimizer is used. Then we switch to SGD (Stochastic Gradient Descent). An example of typical learning curves for training of \textbf{LR~LSTM~500-500} model is given in Fig.~\ref{fig:learn_curves}. 
In the automated tuning process we try to prevent the overfitting by using conventional regularization techniques including stopping the training procedure when the validation perplexity starts to increase.

\subsection{Compression results}

The main results of our research for different compression models are summarized in Table~\ref{results_table}.

\begin{table}[h!]
\scriptsize{
\begin{center}
\caption{Compression results on PTB dataset}\label{results_table}
\renewcommand{\arraystretch}{1.5}
\begin{tabular}{L{12mm}| C{30mm}|c | C{14mm} | C{14mm} | C{15mm} | C{14mm} }
\hline
&
\textbf{ Model}
& \textbf{ \scriptsize{Size, Mb}  } & \textbf{ \scriptsize{No. of param., M}   } & \textbf{ \scriptsize{Test perplexity}   }
& \textbf{ \scriptsize{Avg. word prediction accuracy}   }
& \textbf{ \scriptsize{Inference time, ms}   } \\
\hline
\hline
\multirow{6}{18mm}{PTB  Baselines } 
& \scriptsize{LSTM 200-200} & 18.6  & 4.64  &
117.659 & 0.0085 & 9.63 \\
\cline{2-7}
& \scriptsize{LSTM 300-300} & 29.8  & 7.45  &
91.95 & 0.0109 & 10.24 \\
\cline{2-7}
& \scriptsize{LSTM 400-400} & 42.24 & 10.56  &
86.687 & 0.0115 & 12.4 \\
\cline{2-7}
& \scriptsize{LSTM 500-500} & 56  & 14  &
84.778 & 0.0118 & 14.03 \\
\cline{2-7}
 & \scriptsize{LSTM 650-650} & 79.1  & 19.7  &
82.07 & 0.0122 & 16.13 \\
\cline{2-7}
 & \scriptsize{RNN 650-650} & 67.6  & 16.9  &
124.371 & 0.008 & 15.91  \\
\cline{2-7}
 & \scriptsize{GRU 650-650} & 72.28  & 18.07  &
92.86 & 0.0108 & 16.94 \\
\cline{2-7}
& \scriptsize{LSTM 1500-1500} & \: 264.1  \: & 66.02  &
78.29  & 0.0128 & 45.47\\
\hline
\hline
\multirow{12}{18mm}{Ours} 
& \scriptsize{LSTM 200-200 pruning output layer 90\% w/o additional training} & 5.5 & 0.5 & 149.31 & 0.0067& 9.56 
 \\
\cline{2-7}
& \scriptsize{LSTM 200-200 pruning output layer 90\%  with additional training} & 5.5 & 0.5 &
121.123 & 0.0083 & 9.56 \\
\cline{2-7}
& \scriptsize{LSTM 200-200 quantization (1 byte per number) }  &   4.7  &  4.64 & 
118.232 & 0.0085 & 9.61 \\
\cline{2-7}
& \scriptsize{LR LSTM 200-200} & 3.712  & 0.928  &
136.115 & 0.0073 & 7.83 \\
\cline{2-7}
& \scriptsize{LR LSTM 300-300} & 8.228  & 2.072 &
113.691 & 0.0088 &  8.39 \\
\cline{2-7}
& \scriptsize{LR LSTM 400-400} & 13.12 & 3.28  &
106.623 & 0.0094 & 8.82 \\
\cline{2-7}
& \scriptsize{LR LSTM 500-500} & 14.336  & 3.584  &
97.282 & 0.0103  & 8.95 \\
\cline{2-7}
%\multirow{3}{18mm}{Ours} 
& \scriptsize{LR LSTM 650-650} & 16.8  & 4.2 &
92.885 & 0.0108 & 9.68\\
\cline{2-7}
& \scriptsize{LR RNN 650-650} & 35  & 8.75 &
134.111 & 0.0075 & 11.03 \\
\cline{2-7}
& \scriptsize{LR GRU 650-650} & 12.76  & 3.19 &
111.06 & 0.009& 8.74\\
\cline{2-7}
& \text{ \scriptsize{TT LSTM 600-600} } & 50.4  & 12.6 &
168.639  & 0.0059 & 16.75 \\  
\cline{2-7}
& \text{ \scriptsize{LR LSTM 1500-1500} } & 94.9  &  23.72 &
89.462 & 0.0112 & 15.70  \\ 
\hline
\end{tabular}
\end{center}
}
\end{table} 

As one can see, the use of pruning and quantization let us obtain suitable compression level, sometimes even without quality loss (e.g., see quantization). Unfortunately, those methods are not well-suited to decrease the running time of inference on mobile devices. In fact, the difference in inference time with the baseline \textbf{LSTM~200-200} is not statistically significant, when using the McNemar's test with significance level 5\%.

An important feature of the LR decompositions is high efficiency of operations on mobile devices. The experiments have demonstrated that our model \textbf{LR~LSTM~650-650} compressed with the LR-factorization is even smaller than \textbf{LSTM~200-200}, though the perplexity of the latter model is much worse and this difference in perplexity is statistically significant. 
The inference time of our compressed model \textbf{LR~LSTM~650-650} on mobile phone remains approximately identical to the inference time in the simple \textbf{LSTM~200-200}. In all cases the difference in the model size and the inference time of compressed and base models are statistically significant.

The best obtained result for the TT decomposition (\textbf{TT~LSTM~600-600}) is even worse than \textbf{LSTM~200-200} both in terms of size and perplexity. Hence, we can conclude that the LR-decomposition is the most suitable technique to compress the recurrent cells, because it decreases the memory space and inference time without large degradation in perplexity.

\begin{table}[h]
\scriptsize{
\begin{center}
\caption{State-of-the-art models on PTB dataset}\label{sota_results}
\renewcommand{\arraystretch}{1.5}
\begin{tabular}{C{38mm}|c|C{20mm}|c|C{20mm}}
\hline
\textbf{ Model } & \textbf{ Size, Mb  } & \textbf{ No. of param., M } & \textbf{Test perplexity}
& \textbf{Avg. word prediction accuracy}\\
\hline
\hline
RNN-LDA+KN-5 cache \cite{DBLP:conf/slt/MikolovZ12} & 36  & 9 &
92 & 0.0109\\
\hline
LSTM 650-650 \cite{Zaremba:RNNReg:2014} & 79.1 & 19.7 &
82.7 & 0.0121\\
\hline
Variational LSTM (MC) \cite{DBLP:conf/nips/GalG16} & 80 & 20 &
78.6 & 0.0127\\
\hline
CharCNN \cite{DBLP:conf/aaai/KimJSR16} & 76 & 19 &
78.9 & 0.0127\\
\hline
Variational RHN  \cite{DBLP:conf/icml/ZillySKS17} & 92 & 23 &
65.4 & 0.0153 \\
\hline
AWD-LSTM \cite{DBLP:journals/corr/abs-1708-02182}  &   88  &  22 & 
55.97 & 0.0179 \\
\hline
AWD-LSTM-MoS \cite{DBLP:journals/corr/abs-1711-03953} &   88  &  22 & 
54.44 & 0.0184\\
\hline
TrellisNet-MoS \cite{DBLP:journals/corr/abs-1810-06682} &   136  &  34 & 
54.19 & 0.0185\\
\hline
\hline
LSTM-SparseVD \cite{DBLP:conf/emnlp/ChirkovaLV18} & 13.248  & 3.312 &
109.2 & 0.0092\\
\hline
LSTM-SparseVD-VOC \cite{DBLP:conf/emnlp/ChirkovaLV18} & 6.688 & 1.672 &
120.2 & 0.0083\\
\hline
\end{tabular}
\end{center}
}
\end{table}

In Table~\ref{sota_results}, the state-of-the-art perplexities for language modeling problem are assembled. In addition, we present in the last two rows of this table the best known results (for the PTB dataset) of compressed RNNs using SparseVD method~\cite{DBLP:conf/emnlp/ChirkovaLV18}. Here the number of parameters for the compressed model from the paper~\cite{DBLP:conf/emnlp/ChirkovaLV18} is computed in line with the remaining models as follows. We assume that all computation is performed in matrix form. Therefore, we take all non-zero words and all non-zero LSTM cells (even though there are parameters containing zero). Then the number of parameters is computed according to Eq.~\ref{nparam}.

It has been shown that despite their higher prediction quality, the number of parameters for the best methods is 3-6 times higher the sizes of our compressed models and these results are statistically significant. Moreover, from a practical viewpoint, the average accuracy of the next word prediction is a more interpretable value. One can notice by comparison of Table~\ref{results_table} and Table~\ref{sota_results}, that these accuracies of our compressed models and the state-of-the-art models are rather close to each other. It is important to emphasize that our approach (Fig.~\ref{fig:pipeline}) makes it possible to obtain lower perplexity and model sizes than the existing compression technique~\cite{DBLP:conf/emnlp/ChirkovaLV18}.

\subsection{Last layer decomposition}

In the next experiments we have analyzed the effectiveness of LR and TT decomposition applied to the last (fully-connected) layer. We fix the neural network architecture and change only this layer. We perform a randomized search for each network over the following parameters: internal size of decomposition, TT-ranks, starting learning rate, learning rate schedule, and dropout rate. The best improvements in speed and memory consumption achieved by compressing only the last high-dimensional layer are shown in Table~\ref{results_tt_softmax}.

\begin{table}[ht]
\scriptsize{
\begin{center}
\caption{ LR and TT decomposition of the output layer }
\label{results_tt_softmax}
\renewcommand{\arraystretch}{1.5}
\begin{tabular}{L{18mm} | c|c | C{18mm} |C{20mm} |  C{15mm} }
\hline
&
\textbf{ Model}
& \textbf{ \scriptsize{Size, Mb}  } & \textbf{ \scriptsize{No. of parameters, M }   } &  \textbf{ \scriptsize{No. of output layer parameters, M }   }  & \textbf{ \scriptsize{Test perplexity}   } \\
\hline
\hline
\multirow{3}{18mm}{PTB  Benchmarks }
  &
\scriptsize{LSTM 200-200} & 18.6 & 4.64 & 2.0 &
117.659  \\
\cline{2-6}
& \scriptsize{LSTM 650-650} & 79.1  & 19.7  &
6.5 & 82.07 \\
\cline{2-6}
& \scriptsize{LSTM 1500-1500} & \: 264.1 \: & 66.02 & 15.0 &
78.29 \\
\hline
\multirow{3}{18mm}{LR for Softmax layer} & \scriptsize{ LSTM 200-200} & 12.6  & 3.15  &  0.51&
112.065 \\
\cline{2-6}
& \text{ \scriptsize{LSTM 650-650} } & 57.9 &  14.48 &  1.193 & 84.12  \\ 
\cline{2-6}
& \text{ \scriptsize{LSTM 1500-1500} } & 215.4 &  53.85 &  2.829 & 89.613 \\ 
\hline
\multirow{3}{18mm}{TT for Softmax layer} & \scriptsize{ LSTM 200-200} & 11.8  & 2.95  &  0.304&
116.588 \\
\cline{2-6}
& \text{ \scriptsize{LSTM 600-600} } & 51.12 &  12.8&  1.03 & 88.551  \\ 
\cline{2-6}
& \text{ \scriptsize{LSTM 1500-1500} } & 215.8 &   53.95 &  2.92 & 85.63  \\ 
\hline
\end{tabular}
\end{center}
}
\end{table}

Here, the decomposition of the last layer only reduces the model size in 1.2-1.5 times. Even though in general perplexity is increased, we have succeeded to \textit{decrease} perplexity of the simplest \textbf{LSTM~200-200} model. One can notice that the TT-decomposition is rather promising in this particular task. However, the achieved quality of TT representation is still very unstable and drastically depends on the learning parameters. Hence, usually we have to examine much more configurations to obtain admissible results.

\section{Conclusion}\label{sec:conclusion}

In this paper, we examined several methods of RNNs compression for the language modeling problem. Much attention was paid to the specific problem of high-dimensional output, which is especially crucial in language modeling due to the very large vocabulary size. Such techniques as pruning, quantization, low-rank matrix factorization and Tensor Train decomposition were experimentally compared in terms of their accuracy (perplexity), model size and inference speed on the real mobile device. We tested them across different types of RNNs (LSTM/GRU/RNN) and different model sizes (Table~\ref{results_table}). 

As a result, we formulated the general methodology (Fig.~\ref{fig:pipeline}) for compressing such types of models and make them suitable for implementation in offline mobile applications. Our pipeline is suitable for any such net with LSTM cells and high-dimensional input and output matrices (Table~\ref{results_table}, Table~\ref{results_tt_softmax}).  At the moment of submission, this is one of the first implementations of RNNs (moreover, compressed ones) for mobile GPUs.

It was shown that the main benefit obtained from compression of RNNs by means of LR matrix decomposition in comparison to pruning and quantization lies in the fact that we almost do not lose the speed of matrix multiplication and the memory gain becomes almost equal to the operations gain. In contrast, nowadays, many methods works with sparse matrices, which are able to provide memory gain, but fail with operations gain. Our experimental results on the mobile device confirmed that the LR-compression of the model \textbf{LR~LSTM~650-650} is more efficient in both memory and running-time complexity of the inference. 

Since our approach is studied for recurrent neural nets, certain state-of-the-art models (Table~\ref{sota_results}) that are based on alternative RNN modifications (e.g., Trellis network, RHN, and AWD-LSTM-MOS) of classic architectures have not been tested along with our compression schemes due to their ``fragile'' architecture with many separate hacks applied. Hence, as for the prospective venues of future research, we leave the implementation of compression methods for these complex models, many of which have major modifications in comparison to conventional LSTMs/GRUs and require specialized individual treatment.

\subsubsection*{Acknowledgements.} 
The work of A.V.~Savchenko and D.I.~Ignatov was prepared within the framework of the Basic Research Program at the National Research University Higher School of Economics (HSE) and supported within the framework of a subsidy by the Russian Academic Excellence Project '5-100'. The authors have no conflicts of interest to declare.

The authors would like to thank Dmitriy Polubotko for his valuable help with the experiments on mobile devices.

\section*{References}

\bibliographystyle{model1-num-names}
\bibliography{grbib}

\begin{thebibliography}{47}
\expandafter\ifx\csname natexlab\endcsname\relax\def\natexlab#1{#1}\fi
\providecommand{\bibinfo}[2]{#2}
\ifx\xfnm\relax \def\xfnm[#1]{\unskip,\space#1}\fi
%Type = Book
\bibitem[{Croft and Lafferty(2003)}]{croft2003language}
\bibinfo{author}{W.~B. Croft}, \bibinfo{author}{J.~Lafferty},
  \bibinfo{title}{Language modeling for information retrieval},
  \bibinfo{publisher}{Springer Science \& Business Media},
  \bibinfo{year}{2003}.
%Type = Article
\bibitem[{Deng et~al.(2018)Deng, Zhang, and Shu}]{deng2018feature}
\bibinfo{author}{H.~Deng}, \bibinfo{author}{L.~Zhang},
  \bibinfo{author}{X.~Shu},
\newblock \bibinfo{title}{Feature memory-based deep recurrent neural network
  for language modeling},
\newblock \bibinfo{journal}{Appl. Soft Comput.} \bibinfo{volume}{68}
  (\bibinfo{year}{2018}) \bibinfo{pages}{432--446}.
%Type = Incollection
\bibitem[{Jelinek and Mercer(1980)}]{JelMer80}
\bibinfo{author}{F.~Jelinek}, \bibinfo{author}{R.~L. Mercer},
\newblock \bibinfo{title}{Interpolated estimation of {M}arkov source parameters
  from sparse data},
\newblock in: \bibinfo{editor}{E.~S. Gelsema}, \bibinfo{editor}{L.~N. Kanal}
  (Eds.), \bibinfo{booktitle}{Proceedings, Workshop on Pattern Recognition in
  Practice}, \bibinfo{publisher}{North Holland}, \bibinfo{address}{Amsterdam},
  \bibinfo{year}{1980}, pp. \bibinfo{pages}{381--397}.
%Type = Inproceedings
\bibitem[{Kneser and Ney(1995)}]{Kneser:ibfmlm:1995}
\bibinfo{author}{R.~Kneser}, \bibinfo{author}{H.~Ney},
\newblock \bibinfo{title}{Improved backing-off for m-gram language modeling},
\newblock in: \bibinfo{booktitle}{1995 International Conference on Acoustics,
  Speech, and Signal Processing, {ICASSP} '95, Detroit, Michigan, USA, May
  08-12, 1995}, pp. \bibinfo{pages}{181--184}.
%Type = Book
\bibitem[{Jelinek(1997)}]{Jelinek:SMfSR:1997}
\bibinfo{author}{F.~Jelinek}, \bibinfo{title}{Statistical Methods for Speech
  Recognition}, \bibinfo{publisher}{MIT Press}, \bibinfo{year}{1997}.
%Type = Article
\bibitem[{Hochreiter and Schmidhuber(1997)}]{Schmid:LSTM:1997}
\bibinfo{author}{S.~Hochreiter}, \bibinfo{author}{J.~Schmidhuber},
\newblock \bibinfo{title}{Long short-term memory},
\newblock \bibinfo{journal}{Neural Computation} \bibinfo{volume}{9}
  (\bibinfo{year}{1997}) \bibinfo{pages}{1735--1780}.
%Type = Inproceedings
\bibitem[{Cho et~al.(2014)Cho, van Merrienboer, Bahdanau, and
  Bengio}]{GRU:Cho:2014}
\bibinfo{author}{K.~Cho}, \bibinfo{author}{B.~van Merrienboer},
  \bibinfo{author}{D.~Bahdanau}, \bibinfo{author}{Y.~Bengio},
\newblock \bibinfo{title}{On the properties of neural machine translation:
  Encoder-decoder approaches},
\newblock in: \bibinfo{booktitle}{Proceedings of SSST@EMNLP 2014, Eighth
  Workshop on Syntax, Semantics and Structure in Statistical Translation, Doha,
  Qatar, 25 October 2014}, pp. \bibinfo{pages}{103--111}.
%Type = Inproceedings
\bibitem[{Pham et~al.(2016)Pham, Kruszewski, and
  Boleda}]{DBLP:conf/emnlp/PhamKB16}
\bibinfo{author}{N.~Pham}, \bibinfo{author}{G.~Kruszewski},
  \bibinfo{author}{G.~Boleda},
\newblock \bibinfo{title}{Convolutional neural network language models},
\newblock in: \bibinfo{booktitle}{Proceedings of the 2016 Conference on
  Empirical Methods in Natural Language Processing, {EMNLP} 2016, Austin,
  Texas, USA, November 1-4, 2016}, pp. \bibinfo{pages}{1153--1162}.
%Type = Inproceedings
\bibitem[{Kim et~al.(2016)Kim, Jernite, Sontag, and
  Rush}]{DBLP:conf/aaai/KimJSR16}
\bibinfo{author}{Y.~Kim}, \bibinfo{author}{Y.~Jernite},
  \bibinfo{author}{D.~Sontag}, \bibinfo{author}{A.~M. Rush},
\newblock \bibinfo{title}{Character-aware neural language models},
\newblock in: \bibinfo{booktitle}{Proceedings of the Thirtieth {AAAI}
  Conference on Artificial Intelligence, February 12-17, 2016, Phoenix,
  Arizona, {USA.}}, pp. \bibinfo{pages}{2741--2749}.
%Type = Article
\bibitem[{Devlin et~al.(2018)Devlin, Chang, Lee, and
  Toutanova}]{DBLP:journals/corr/abs-1810-04805}
\bibinfo{author}{J.~Devlin}, \bibinfo{author}{M.~Chang},
  \bibinfo{author}{K.~Lee}, \bibinfo{author}{K.~Toutanova},
\newblock \bibinfo{title}{{BERT:} pre-training of deep bidirectional
  transformers for language understanding},
\newblock \bibinfo{journal}{CoRR} \bibinfo{volume}{abs/1810.04805}
  (\bibinfo{year}{2018}).
%Type = Inproceedings
\bibitem[{Vaswani et~al.(2017)Vaswani, Shazeer, Parmar, Uszkoreit, Jones,
  Gomez, Kaiser, and Polosukhin}]{DBLP:conf/nips/VaswaniSPUJGKP17}
\bibinfo{author}{A.~Vaswani}, \bibinfo{author}{N.~Shazeer},
  \bibinfo{author}{N.~Parmar}, \bibinfo{author}{J.~Uszkoreit},
  \bibinfo{author}{L.~Jones}, \bibinfo{author}{A.~N. Gomez},
  \bibinfo{author}{L.~Kaiser}, \bibinfo{author}{I.~Polosukhin},
\newblock \bibinfo{title}{Attention is all you need},
\newblock in: \bibinfo{booktitle}{Advances in Neural Information Processing
  Systems 30: Annual Conference on Neural Information Processing Systems 2017,
  4-9 December 2017, Long Beach, CA, {USA}}, pp. \bibinfo{pages}{6000--6010}.
%Type = Article
\bibitem[{Cheok et~al.(2008)Cheok, Zhang, and Siong}]{cheok2008efficient}
\bibinfo{author}{A.~D. Cheok}, \bibinfo{author}{J.~Zhang},
  \bibinfo{author}{C.~E. Siong},
\newblock \bibinfo{title}{Efficient mobile phone chinese optical character
  recognition systems by use of heuristic fuzzy rules and bigram markov
  language models},
\newblock \bibinfo{journal}{Appl. Soft Comput.} \bibinfo{volume}{8}
  (\bibinfo{year}{2008}) \bibinfo{pages}{1005--1017}.
%Type = Inproceedings
\bibitem[{Bengio and Senecal(2003)}]{Bengio+Senecal-2003}
\bibinfo{author}{Y.~Bengio}, \bibinfo{author}{J.~Senecal},
\newblock \bibinfo{title}{Quick training of probabilistic neural nets by
  importance sampling},
\newblock in: \bibinfo{booktitle}{Proceedings of the Ninth International
  Workshop on Artificial Intelligence and Statistics, {AISTATS} 2003, Key West,
  Florida, USA, January 3-6, 2003}.
%Type = Inproceedings
\bibitem[{Grachev et~al.(2017)Grachev, Ignatov, and Savchenko}]{Grachev1}
\bibinfo{author}{A.~M. Grachev}, \bibinfo{author}{D.~I. Ignatov},
  \bibinfo{author}{A.~V. Savchenko},
\newblock \bibinfo{title}{Neural networks compression for language modeling},
\newblock in: \bibinfo{booktitle}{Pattern Recognition and Machine Intelligence
  - 7th International Conference, PReMI 2017, Kolkata, India, December 5-8,
  2017, Proceedings}, pp. \bibinfo{pages}{351--357}.
%Type = Article
\bibitem[{Zaremba et~al.(2014)Zaremba, Sutskever, and
  Vinyals}]{Zaremba:RNNReg:2014}
\bibinfo{author}{W.~Zaremba}, \bibinfo{author}{I.~Sutskever},
  \bibinfo{author}{O.~Vinyals},
\newblock \bibinfo{title}{Recurrent neural network regularization},
\newblock \bibinfo{journal}{CoRR} \bibinfo{volume}{abs/1409.2329}
  (\bibinfo{year}{2014}).
%Type = Inproceedings
\bibitem[{Han et~al.(2015{\natexlab{a}})Han, Pool, Tran, and Dally}]{pruning1}
\bibinfo{author}{S.~Han}, \bibinfo{author}{J.~Pool}, \bibinfo{author}{J.~Tran},
  \bibinfo{author}{W.~J. Dally},
\newblock \bibinfo{title}{Learning both weights and connections for efficient
  neural network},
\newblock in: \bibinfo{booktitle}{Advances in Neural Information Processing
  Systems 28: Annual Conference on Neural Information Processing Systems 2015,
  December 7-12, 2015, Montreal, Quebec, Canada}, pp.
  \bibinfo{pages}{1135--1143}.
%Type = Article
\bibitem[{Han et~al.(2015{\natexlab{b}})Han, Mao, and
  Dally}]{Han:DeepCompressing:2016}
\bibinfo{author}{S.~Han}, \bibinfo{author}{H.~Mao}, \bibinfo{author}{W.~J.
  Dally},
\newblock \bibinfo{title}{Deep compression: Compressing deep neural network
  with pruning, trained quantization and huffman coding},
\newblock \bibinfo{journal}{CoRR} \bibinfo{volume}{abs/1510.00149}
  (\bibinfo{year}{2015}{\natexlab{b}}).
%Type = Incollection
\bibitem[{Kingma et~al.(2015)Kingma, Salimans, and Welling}]{vardropout}
\bibinfo{author}{D.~P. Kingma}, \bibinfo{author}{T.~Salimans},
  \bibinfo{author}{M.~Welling},
\newblock \bibinfo{title}{Variational dropout and the local reparameterization
  trick},
\newblock in: \bibinfo{editor}{C.~Cortes}, \bibinfo{editor}{N.~D. Lawrence},
  \bibinfo{editor}{D.~D. Lee}, \bibinfo{editor}{M.~Sugiyama},
  \bibinfo{editor}{R.~Garnett} (Eds.), \bibinfo{booktitle}{Advances in Neural
  Information Processing Systems 28}, \bibinfo{publisher}{Curran Associates,
  Inc.}, \bibinfo{year}{2015}, pp. \bibinfo{pages}{2575--2583}.
%Type = Inproceedings
\bibitem[{Molchanov et~al.(2017)Molchanov, Ashukha, and
  Vetrov}]{bayescomperssion1}
\bibinfo{author}{D.~Molchanov}, \bibinfo{author}{A.~Ashukha},
  \bibinfo{author}{D.~P. Vetrov},
\newblock \bibinfo{title}{Variational dropout sparsifies deep neural networks},
\newblock in: \bibinfo{booktitle}{Proceedings of the 34th International
  Conference on Machine Learning, {ICML} 2017, Sydney, NSW, Australia, 6-11
  August 2017}, pp. \bibinfo{pages}{2498--2507}.
%Type = Inproceedings
\bibitem[{Neklyudov et~al.(2017)Neklyudov, Molchanov, Ashukha, and
  Vetrov}]{bayescomperssion2}
\bibinfo{author}{K.~Neklyudov}, \bibinfo{author}{D.~Molchanov},
  \bibinfo{author}{A.~Ashukha}, \bibinfo{author}{D.~P. Vetrov},
\newblock \bibinfo{title}{Structured bayesian pruning via log-normal
  multiplicative noise},
\newblock in: \bibinfo{booktitle}{Advances in Neural Information Processing
  Systems 30: Annual Conference on Neural Information Processing Systems 2017,
  4-9 December 2017, Long Beach, CA, {USA}}, pp. \bibinfo{pages}{6778--6787}.
%Type = Article
\bibitem[{Lobacheva et~al.(2017)Lobacheva, Chirkova, and
  Vetrov}]{bayescomperssion3}
\bibinfo{author}{E.~Lobacheva}, \bibinfo{author}{N.~Chirkova},
  \bibinfo{author}{D.~Vetrov},
\newblock \bibinfo{title}{Bayesian sparsification of recurrent neural
  networks},
\newblock \bibinfo{journal}{arXiv preprint arXiv:1708.00077}
  (\bibinfo{year}{2017}).
%Type = Inproceedings
\bibitem[{Arjovsky et~al.(2016)Arjovsky, Shah, and Bengio}]{Bengio:URNN:2016}
\bibinfo{author}{M.~Arjovsky}, \bibinfo{author}{A.~Shah},
  \bibinfo{author}{Y.~Bengio},
\newblock \bibinfo{title}{Unitary evolution recurrent neural networks},
\newblock in: \bibinfo{booktitle}{Proceedings of the 33nd International
  Conference on Machine Learning, {ICML} 2016, New York City, NY, USA, June
  19-24, 2016}, pp. \bibinfo{pages}{1120--1128}.
%Type = Inproceedings
\bibitem[{Yang et~al.(2015)Yang, Moczulski, Denil, de~Freitas, Smola, Song, and
  Wang}]{Yang:2015}
\bibinfo{author}{Z.~Yang}, \bibinfo{author}{M.~Moczulski},
  \bibinfo{author}{M.~Denil}, \bibinfo{author}{N.~de~Freitas},
  \bibinfo{author}{A.~J. Smola}, \bibinfo{author}{L.~Song},
  \bibinfo{author}{Z.~Wang},
\newblock \bibinfo{title}{Deep fried convnets},
\newblock in: \bibinfo{booktitle}{2015 {IEEE} International Conference on
  Computer Vision, {ICCV} 2015, Santiago, Chile, December 7-13, 2015}, pp.
  \bibinfo{pages}{1476--1483}.
%Type = Inproceedings
\bibitem[{Novikov et~al.(2015)Novikov, Podoprikhin, Osokin, and
  Vetrov}]{novikov15tensornet}
\bibinfo{author}{A.~Novikov}, \bibinfo{author}{D.~Podoprikhin},
  \bibinfo{author}{A.~Osokin}, \bibinfo{author}{D.~P. Vetrov},
\newblock \bibinfo{title}{Tensorizing neural networks},
\newblock in: \bibinfo{booktitle}{Advances in Neural Information Processing
  Systems 28: Annual Conference on Neural Information Processing Systems 2015,
  December 7-12, 2015, Montreal, Quebec, Canada}, pp.
  \bibinfo{pages}{442--450}.
%Type = Article
\bibitem[{Garipov et~al.(2016)Garipov, Podoprikhin, Novikov, and
  Vetrov}]{garipov16ttconv}
\bibinfo{author}{T.~Garipov}, \bibinfo{author}{D.~Podoprikhin},
  \bibinfo{author}{A.~Novikov}, \bibinfo{author}{D.~P. Vetrov},
\newblock \bibinfo{title}{Ultimate tensorization: compressing convolutional and
  {FC} layers alike},
\newblock \bibinfo{journal}{CoRR} \bibinfo{volume}{abs/1611.03214}
  (\bibinfo{year}{2016}).
%Type = Inproceedings
\bibitem[{Tjandra et~al.(2017)Tjandra, Sakti, and Nakamura}]{TTRNN}
\bibinfo{author}{A.~Tjandra}, \bibinfo{author}{S.~Sakti},
  \bibinfo{author}{S.~Nakamura},
\newblock \bibinfo{title}{Compressing recurrent neural network with tensor
  train},
\newblock in: \bibinfo{booktitle}{2017 International Joint Conference on Neural
  Networks, {IJCNN} 2017, Anchorage, AK, USA, May 14-19, 2017}, pp.
  \bibinfo{pages}{4451--4458}.
%Type = Article
\bibitem[{Yu et~al.(2017)Yu, Zheng, Anandkumar, and Yue}]{yu2017long}
\bibinfo{author}{R.~Yu}, \bibinfo{author}{S.~Zheng},
  \bibinfo{author}{A.~Anandkumar}, \bibinfo{author}{Y.~Yue},
\newblock \bibinfo{title}{Long-term forecasting using tensor-train rnns},
\newblock \bibinfo{journal}{CoRR} \bibinfo{volume}{abs/1711.00073}
  (\bibinfo{year}{2017}).
%Type = Inproceedings
\bibitem[{Lu et~al.(2016)Lu, Sindhwani, and Sainath}]{Lu:LR:2016}
\bibinfo{author}{Z.~Lu}, \bibinfo{author}{V.~Sindhwani}, \bibinfo{author}{T.~N.
  Sainath},
\newblock \bibinfo{title}{Learning compact recurrent neural networks},
\newblock in: \bibinfo{booktitle}{2016 {IEEE} International Conference on
  Acoustics, Speech and Signal Processing, {ICASSP} 2016, Shanghai, China,
  March 20-25, 2016}, pp. \bibinfo{pages}{5960--5964}.
%Type = Inproceedings
\bibitem[{Prabhavalkar et~al.(2016)Prabhavalkar, Alsharif, Bruguier, and
  McGraw}]{DBLP:conf/icassp/PrabhavalkarABM16}
\bibinfo{author}{R.~Prabhavalkar}, \bibinfo{author}{O.~Alsharif},
  \bibinfo{author}{A.~Bruguier}, \bibinfo{author}{I.~McGraw},
\newblock \bibinfo{title}{On the compression of recurrent neural networks with
  an application to {LVCSR} acoustic modeling for embedded speech recognition},
\newblock in: \bibinfo{booktitle}{2016 {IEEE} International Conference on
  Acoustics, Speech and Signal Processing, {ICASSP} 2016, Shanghai, China,
  March 20-25, 2016}, pp. \bibinfo{pages}{5970--5974}.
%Type = Article
\bibitem[{Acharya et~al.(2018)Acharya, Goel, Metallinou, and
  Dhillon}]{DBLP:journals/corr/abs-1811-00641}
\bibinfo{author}{A.~Acharya}, \bibinfo{author}{R.~Goel},
  \bibinfo{author}{A.~Metallinou}, \bibinfo{author}{I.~S. Dhillon},
\newblock \bibinfo{title}{Online embedding compression for text classification
  using low rank matrix factorization},
\newblock \bibinfo{journal}{CoRR} \bibinfo{volume}{abs/1811.00641}
  (\bibinfo{year}{2018}).
%Type = Inproceedings
\bibitem[{G{\"{u}}l{\c{c}}ehre et~al.(2016)G{\"{u}}l{\c{c}}ehre, Ahn,
  Nallapati, Zhou, and Bengio}]{GulcehreANZB16}
\bibinfo{author}{{\c{C}}.~G{\"{u}}l{\c{c}}ehre}, \bibinfo{author}{S.~Ahn},
  \bibinfo{author}{R.~Nallapati}, \bibinfo{author}{B.~Zhou},
  \bibinfo{author}{Y.~Bengio},
\newblock \bibinfo{title}{Pointing the unknown words},
\newblock in: \bibinfo{booktitle}{Proceedings of the 54th Annual Meeting of the
  Association for Computational Linguistics, {ACL} 2016, August 7-12, 2016,
  Berlin, Germany, Volume 1: Long Papers}.
%Type = Inproceedings
\bibitem[{Morin and Bengio(2005)}]{Morin05hierarchicalprobabilistic}
\bibinfo{author}{F.~Morin}, \bibinfo{author}{Y.~Bengio},
\newblock \bibinfo{title}{Hierarchical probabilistic neural network language
  model},
\newblock in: \bibinfo{booktitle}{Proceedings of the Tenth International
  Workshop on Artificial Intelligence and Statistics, {AISTATS} 2005,
  Bridgetown, Barbados, January 6-8, 2005}.
%Type = Inproceedings
\bibitem[{Chirkova et~al.(2018)Chirkova, Lobacheva, and
  Vetrov}]{DBLP:conf/emnlp/ChirkovaLV18}
\bibinfo{author}{N.~Chirkova}, \bibinfo{author}{E.~Lobacheva},
  \bibinfo{author}{D.~P. Vetrov},
\newblock \bibinfo{title}{Bayesian compression for natural language
  processing},
\newblock in: \bibinfo{booktitle}{Proceedings of the 2018 Conference on
  Empirical Methods in Natural Language Processing, Brussels, Belgium, October
  31 - November 4, 2018}, pp. \bibinfo{pages}{2910--2915}.
%Type = Inproceedings
\bibitem[{Mikolov et~al.(2010)Mikolov, Karafi{\'{a}}t, Burget, Cernock{\'{y}},
  and Khudanpur}]{DBLP:conf/interspeech/MikolovKBCK10}
\bibinfo{author}{T.~Mikolov}, \bibinfo{author}{M.~Karafi{\'{a}}t},
  \bibinfo{author}{L.~Burget}, \bibinfo{author}{J.~Cernock{\'{y}}},
  \bibinfo{author}{S.~Khudanpur},
\newblock \bibinfo{title}{Recurrent neural network based language model},
\newblock in: \bibinfo{booktitle}{{INTERSPEECH} 2010, 11th Annual Conference of
  the International Speech Communication Association, Makuhari, Chiba, Japan,
  September 26-30, 2010}, pp. \bibinfo{pages}{1045--1048}.
%Type = Article
\bibitem[{Bengio et~al.(2003)Bengio, Ducharme, Vincent, and
  Janvin}]{Bengio03aneural}
\bibinfo{author}{Y.~Bengio}, \bibinfo{author}{R.~Ducharme},
  \bibinfo{author}{P.~Vincent}, \bibinfo{author}{C.~Janvin},
\newblock \bibinfo{title}{A neural probabilistic language model},
\newblock \bibinfo{journal}{Journal of Machine Learning Research}
  \bibinfo{volume}{3} (\bibinfo{year}{2003}) \bibinfo{pages}{1137--1155}.
%Type = Phdthesis
\bibitem[{Mikolov(2012)}]{Mikolov:2007}
\bibinfo{author}{T.~Mikolov}, \bibinfo{title}{Statistical Language Models Based
  on Neural Networks}, Ph.D. thesis, Brno University of Technology,
  \bibinfo{year}{2012}.
%Type = Article
\bibitem[{Hochreiter et~al.(2001)Hochreiter, Bengio, Frasconi, and
  Schmidhuber}]{Gradient_flow:Hochreiter:2001}
\bibinfo{author}{S.~Hochreiter}, \bibinfo{author}{Y.~Bengio},
  \bibinfo{author}{P.~Frasconi}, \bibinfo{author}{J.~Schmidhuber},
\newblock \bibinfo{title}{Gradient flow in recurrent nets: the difficulty of
  learning long-term dependencies},
\newblock \bibinfo{journal}{S. C. Kremer and J. F. Kolen, eds. A Field Guide to
  Dynamical Recurrent Neural Networks}  (\bibinfo{year}{2001}).
%Type = Article
\bibitem[{Inan et~al.(2016)Inan, Khosravi, and
  Socher}]{DBLP:journals/corr/InanKS16}
\bibinfo{author}{H.~Inan}, \bibinfo{author}{K.~Khosravi},
  \bibinfo{author}{R.~Socher},
\newblock \bibinfo{title}{Tying word vectors and word classifiers: {A} loss
  framework for language modeling},
\newblock \bibinfo{journal}{CoRR} \bibinfo{volume}{abs/1611.01462}
  (\bibinfo{year}{2016}).
%Type = Inproceedings
\bibitem[{Press and Wolf(2017)}]{E17-2025}
\bibinfo{author}{O.~Press}, \bibinfo{author}{L.~Wolf},
\newblock \bibinfo{title}{Using the output embedding to improve language
  models},
\newblock in: \bibinfo{booktitle}{Proceedings of the 15th Conference of the
  European Chapter of the Association for Computational Linguistics, {EACL}
  2017, Valencia, Spain, April 3-7, 2017, Volume 2: Short Papers}, pp.
  \bibinfo{pages}{157--163}.
%Type = Article
\bibitem[{Rassadin and Savchenko(2017)}]{Image:Svachenko:2017}
\bibinfo{author}{A.~G. Rassadin}, \bibinfo{author}{A.~V. Savchenko},
\newblock \bibinfo{title}{Deep neural networks performance optimization in
  image recognition},
\newblock \bibinfo{journal}{Proceedings of the 3rd International Conference on
  Information Technologies and Nanotechnologies (ITNT)}
  (\bibinfo{year}{2017}).
%Type = Article
\bibitem[{Oseledets(2011)}]{TT:Oseledets:11}
\bibinfo{author}{I.~V. Oseledets},
\newblock \bibinfo{title}{Tensor-train decomposition},
\newblock \bibinfo{journal}{{SIAM} J. Scientific Computing}
  \bibinfo{volume}{33} (\bibinfo{year}{2011}) \bibinfo{pages}{2295--2317}.
%Type = Inproceedings
\bibitem[{Mikolov and Zweig(2012)}]{DBLP:conf/slt/MikolovZ12}
\bibinfo{author}{T.~Mikolov}, \bibinfo{author}{G.~Zweig},
\newblock \bibinfo{title}{Context dependent recurrent neural network language
  model},
\newblock in: \bibinfo{booktitle}{2012 {IEEE} Spoken Language Technology
  Workshop (SLT), Miami, FL, USA, December 2-5, 2012}, pp.
  \bibinfo{pages}{234--239}.
%Type = Inproceedings
\bibitem[{Gal and Ghahramani(2016)}]{DBLP:conf/nips/GalG16}
\bibinfo{author}{Y.~Gal}, \bibinfo{author}{Z.~Ghahramani},
\newblock \bibinfo{title}{A theoretically grounded application of dropout in
  recurrent neural networks},
\newblock in: \bibinfo{booktitle}{Advances in Neural Information Processing
  Systems 29: Annual Conference on Neural Information Processing Systems 2016,
  December 5-10, 2016, Barcelona, Spain}, pp. \bibinfo{pages}{1019--1027}.
%Type = Inproceedings
\bibitem[{Zilly et~al.(2017)Zilly, Srivastava, Koutn{\'{\i}}k, and
  Schmidhuber}]{DBLP:conf/icml/ZillySKS17}
\bibinfo{author}{J.~G. Zilly}, \bibinfo{author}{R.~K. Srivastava},
  \bibinfo{author}{J.~Koutn{\'{\i}}k}, \bibinfo{author}{J.~Schmidhuber},
\newblock \bibinfo{title}{Recurrent highway networks},
\newblock in: \bibinfo{booktitle}{Proceedings of the 34th International
  Conference on Machine Learning, {ICML} 2017, Sydney, NSW, Australia, 6-11
  August 2017}, pp. \bibinfo{pages}{4189--4198}.
%Type = Article
\bibitem[{Merity et~al.(2017)Merity, Keskar, and
  Socher}]{DBLP:journals/corr/abs-1708-02182}
\bibinfo{author}{S.~Merity}, \bibinfo{author}{N.~S. Keskar},
  \bibinfo{author}{R.~Socher},
\newblock \bibinfo{title}{Regularizing and optimizing {LSTM} language models},
\newblock \bibinfo{journal}{CoRR} \bibinfo{volume}{abs/1708.02182}
  (\bibinfo{year}{2017}).
%Type = Article
\bibitem[{Yang et~al.(2017)Yang, Dai, Salakhutdinov, and
  Cohen}]{DBLP:journals/corr/abs-1711-03953}
\bibinfo{author}{Z.~Yang}, \bibinfo{author}{Z.~Dai},
  \bibinfo{author}{R.~Salakhutdinov}, \bibinfo{author}{W.~W. Cohen},
\newblock \bibinfo{title}{Breaking the softmax bottleneck: {A} high-rank {RNN}
  language model},
\newblock \bibinfo{journal}{CoRR} \bibinfo{volume}{abs/1711.03953}
  (\bibinfo{year}{2017}).
%Type = Article
\bibitem[{Bai et~al.(2018)Bai, Kolter, and
  Koltun}]{DBLP:journals/corr/abs-1810-06682}
\bibinfo{author}{S.~Bai}, \bibinfo{author}{J.~Z. Kolter},
  \bibinfo{author}{V.~Koltun},
\newblock \bibinfo{title}{Trellis networks for sequence modeling},
\newblock \bibinfo{journal}{CoRR} \bibinfo{volume}{abs/1810.06682}
  (\bibinfo{year}{2018}).

\end{thebibliography}

\end{document}